\documentclass[runningheads]{llncs}
\usepackage{graphicx}
\usepackage{amsmath,amssymb} 
\usepackage{color}
\usepackage{multirow}
\usepackage{booktabs}
\usepackage{hyperref}

\begin{document}
\pagestyle{headings}
\mainmatter

\def\ACCV20SubNumber{287}  

\title{Sequential View Synthesis with Transformer} 

\author{Phong Nguyen-Ha\inst{1}\orcidID{0000-0002-9678-0886} \and
Lam Huynh\inst{1}\orcidID{0000-0002-8311-1288} \and \\
Esa Rahtu\inst{2}\orcidID{0000-0001-8767-0864} \and
Janne Heikkil\"a\inst{1}\orcidID{0000-0003-0073-0866} }
\authorrunning{P. Nguyen et al.}
\institute{Center for Machine Vision and Signal Analysis, University of Oulu, Finland 
\email{\{phong.nguyen,lam.huynh,janne.heikkila\}@oulu.fi}
\and
Computer Vision Group, Tampere University, Finland \\
\email{esa.rahtu@tuni.fi} } 

\maketitle

\begin{abstract}
This paper addresses the problem of novel view synthesis by means of neural rendering, where we are interested in predicting the novel view at an arbitrary camera pose based on a given set of input images from other viewpoints. Using the known query pose and input poses, we create an ordered set of observations that leads to the target view. Thus, the problem of single novel view synthesis is reformulated as a sequential view prediction task. In this paper, the proposed Transformer-based Generative Query Network (T-GQN) extends the neural-rendering methods by adding two new concepts. First, we use multi-view attention learning between context images to obtain multiple implicit scene representations. Second, we introduce a sequential rendering decoder to predict an image sequence, including the target view, based on the learned representations. Finally, we evaluate our model on various challenging datasets and demonstrate that our model not only gives consistent predictions but also doesn't require any retraining for finetuning.
\end{abstract}


\section{Introduction}\label{intro}

View synthesis aims to create novel views of an object or a scene from a perspective of a virtual camera based on a set of reference images. It has been an active field of research already for several decades in computer vision and computer graphics due to its various application areas including free-viewpoint television, virtual and augmented reality, and telepresence \cite{Intro_1,Intro_2,Intro_3,VR_1,VR_2}. 

Conventionally, the view synthesis problem has been addressed by using image-based or geometry-based approaches \cite{Intro_1}. In pure image-based rendering, the novel view is warped from a densely sampled set of reference images or the light field without exploiting any geometric information, which obviously requires a large amount of image data and limits the new viewpoints to a relatively small range. In geometry-based rendering, the novel view is generated using a 3D model that has been first created from the reference views using multi-view stereo or some other image-based modeling techniques. This allows for larger baselines between the views, but also sets high requirements to the quality of the 3D model. Between these two extremes, free viewpoint depth-image-based rendering (DIBR) uses depth maps associated to the reference views enabling 3D image warping to synthesize the novel view \cite{Intro_3}. In practice, all these approaches tend to produce notable artifacts due to missing or inaccurate data, which reduced the quality of the rendered image. 

\begin{figure}[ht]
    \centering
    \includegraphics[width=\textwidth]{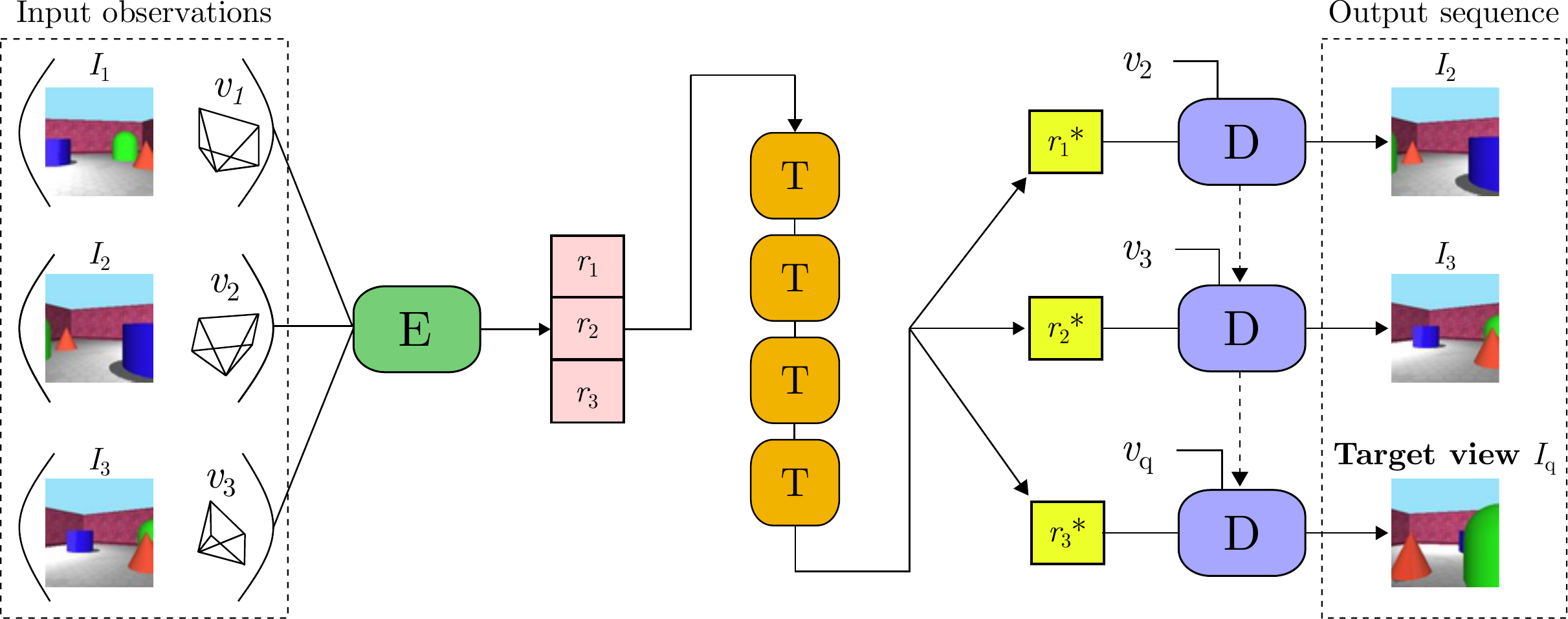}
    \caption{Overview of our proposed Transformer-based Generative Query Network (T-GQN). Given a set of input images and camera poses $\{(I_n,v_n)\}$, where $n=1,\ldots,N$, we obtain a set of $N$ scene representations by passing them one-by-one through the encoder (E) network. We use a stack of transformer encoder (T) to perform the multi-view attention learning between these representation. Finally, we use each these attention representation $r_n^*$ to sequentially render the output sequence using the decoder (D) network. The dashed arrow indicates that we initialize states of the decoder network using the computed states of the previous step.}
    \label{overall_structure}
\end{figure} 

Recently, researchers have adopted deep learning techniques to overcome the inherent limitation of the conventional approaches. This paper focuses on a family of neural rendering methods \cite{GQN,GQN_reversed,CGQN,GAQN,EGQN} that infer the underlying 3D scene structure and faithfully produces the target view even at a distant query pose. These methods use an aggregate function to represent the entire 3D scene as a single implicit representation. Although they manage to successfully render the target view, a large amount of data is required to train, which in turn, takes a long time to converge. We argue that because these methods focus on synthesizing only a single target image, they are inefficient when rendering target images for distant query poses that are subject to strong geometric transformations and occlusions with respect to the reference images.

In this paper, we introduce a Transformer-based Generative Query Network (T-GQN) to address the problem of novel view synthesis in a sequential manner. We train an end-to-end model that sequentially renders a positionally ordered set of nearby-views and then predicts the target view at the final rendering step (as shown in Fig. \ref{overall_structure}). We claim that if the model is able to render the nearby-views accurately, it will be also capable of predicting the target view correctly. Since we do not have the nearby-views of the target pose, we train our model to render the input views. Moreover, instead of rendering the viewpoints using a single implicit scene representation, we use multi-view attention learning based on Transformer Encoder \cite{Transformer_1} to learn multiple scene representations. At each rendering step, we modify the decoder network of Generative Query Network (GQN) \cite{GQN} to share its states through rendering steps.
To summarize, our key contributions are as follows:
\begin{enumerate}
  \item  We reformulate the problem of single view synthesis into sequential view synthesis.\\
  \item Our proposed T-GQN introduces two novel concepts: multi-view attention learning via the Transformer Encoder and a sequential rendering decoder. Our model extends the previously proposed GQN by sequentially rendering a pose-ordered set of novel views.    \\
  \item We demonstrate that our proposed framework achieves state-of-the-art performance on challenging view synthesis datasets. Moreover, our method reaches convergence faster and requires less computational resource to train. 
\end{enumerate}
The source code and the model are available at \url{https://github.com/phongnhhn92/TransformerGQN} .

\section{Related Work}\label{RelatedWorks}
The literature related to view synthesis is extensive, and to limit the scope, we focus on a few deep learning based methods in this section that are most relevant to our method. We suggest a state-of-the-art report for an extensive review \cite{SOTAreport}. 


Many early solutions use regression to derive the pixel colors of the target view directly from the input images \cite{previous_3,mono_viewcontrol,previous_1,AppearanceFlow,previous_2}. In \cite{previous_1}, Tatarchenko et al. maps an image of a scene to an RGB-D image from an unknown viewpoint with an auto-encoder architecture and train their model using supervised learning. Instead of synthesizing pixels from scratch, other works explores using CNNs to predict appearance flow \cite{AppearanceFlow}. Later work by Sun et al. \cite{previous_2} presents a self-confidence aggregation mechanism to integrate both predicted appearance flow and pixel hallucination to achieve contractually consistent results. Although, those method manage to render plausible novel views, their results are limited to a scene with a single object or a slight change in viewpoints in the autonomous driving situation. \par

A great amount of effort has been dedicated to incorporate geometric information to the model. For example, \cite{Deep_stereo,StereoMagnification,lightfield1,MPIImages,extremeview,single_view_mpi} apply deep learning techniques to leverage geometry cues and learn to predict the novel view. The deep learning based-light field camera view interpolation \cite{lightfield1,LLFF} use a deep network to predict depth separately for every novel view. 
Another line of work \cite{Deep_stereo,StereoMagnification,MPIImages} cleverly extract a Multiplane image representation of the scene. This representation offers regularization that allows for an impressive stereo baseline extrapolation. Recently, Choi et al.\cite{extremeview} use deep neural network to estimate a depth probability volume, rather than just a single depth value for each pixel of the novel view. Even though, they show promising results but they are limited to synthesizing a middle view among source images or a magnified view from a single input. In contrast, our proposed framework focuses on arbitrary target views and is able to learn from source images that vary in length.\par

Recent progresses in geometric deep learning proposes to represent a scene as a voxel-grid to enforce the 3D structure. These methods \cite{holoGAN,deep_voxels,neural_volumes} use 3D convolution layers to learn 3D spatial transformations from the input views to the novel view and then apply GAN training \cite{GAN} to enhance quality of the output image. Recently, Sitzman et al. \cite{srn} presents a continuous, 3D-structure-aware scene representation that encodes both geometry and appearance. Their work learns a mapping from world coordinates to feature representation of local scene properties. Another contribution of this paper is that authors empirically demonstrated that their method is able to show generalization across scenes for classes of single objects using an additional MLP network such as HyperNetwork \cite{hypernetwork}. In our paper, we evaluate our method on the novel view synthesis dataset which contains 2 millions scenes each (multiple combinations of  objects types, colors, lightning positions). We argue that training their method to generalize to a large number of instances would be computationally expensive. In fact, this is an open research problem that has not yet been solved.

Recent \textbf{neural rendering} methods have introduced a generative model that understands the underlying 3D scene structure and faithfully produces the target view at the distant query pose \cite{nerf,Plenoptic,equivariant,differentiable}. Generative Query Network (GQN) \cite{GQN} and its variant \cite{GQN_reversed,CGQN,GAQN} are incorporating all input observation (images and poses) into a single implicit 3D scene representation to generate the target view. This aggregated representation contains all necessary information (e.g. object identities, positions, colors, scene layout) to make accurate image predictions. Moreover, these methods corresponds to a  special case of Neural Processes \cite{NP1,NP2}. In this paper, we argue that generating the target view using such compact representation leads to poor predictions and slow training convergence. \par

\section{Approach}\label{Methods}

In this section, we first provide the reader with a brief background of the Generative Query Network (GQN). Then we convert the problem of \textbf{single view synthesis} to the problem of \textbf{sequential view synthesis} by introducing our Transformer-based Generative Query Network (T-GQN) that extends the current GQN architecture with two novel building blocks: multi-view attention learning via Transformer Encoder and sequential rendering decoder.

\subsection{Generative Query Network}\label{Background}
Given the observations that include $N$ images $I_n \in I$ and their corresponding camera poses $v_n \in V$, GQN \cite{GQN} solves the \textbf{single view synthesis} problem by using a encoder-decoder neural network to predict the target image $I_q$ at an arbitrary query pose $v_q$. 

First, the encoder is a feed-forward neural network that takes $N$ observations as input and produce a single implicit scene representation $R=\sum_{n=1}^{N}r_n$ by performing a element-wise sum of $N$ encoded scene representation $r_n$. The decoder then takes $R$ and $v_q$ as an input and predicts the new view $I_q^\prime$ from that viewpoint. The decoder network is a conditional latent variable model DRAW \cite{draw1,draw2} which includes $M$ pairs of Generation and Inference convolutional LSTM networks. At each generation step, the hidden state of the Generation and Inference LSTM core is utilized to approximate the prior $\pi$ and posterior distribution $q$. Since the target view $I_q$ is fed into the Inference sub-network, minimizing the Kullback-Leibler (KL) distance between $\pi$ and $q$ would help the Generation sub-network to produce an accurate result. Both the encoder and decoder networks are trained jointly to minimize the ELBO loss $\mathcal{L}_{GQN}$ function:
\begin{equation}
\label{eqn_GQN}
\mathcal{L}_{GQN} = \bigg[
    -\ln\mathcal{N}(I_q|I_q^\prime)+\sum_{m=1}^{M} KL\Big[\mathcal{N}(q_m) || \mathcal{N}({\pi}_m)\Big]
\bigg]
\end{equation}
In the next section, we will describe how we reformulate the problem of \textbf{single view synthesis} to the problem of \textbf{sequential view synthesis} to address this issue.

\subsection{Sequential view synthesis}
\label{seq_synthesis}
\begin{figure}[!t]
    \centering
    \includegraphics[width=\textwidth]{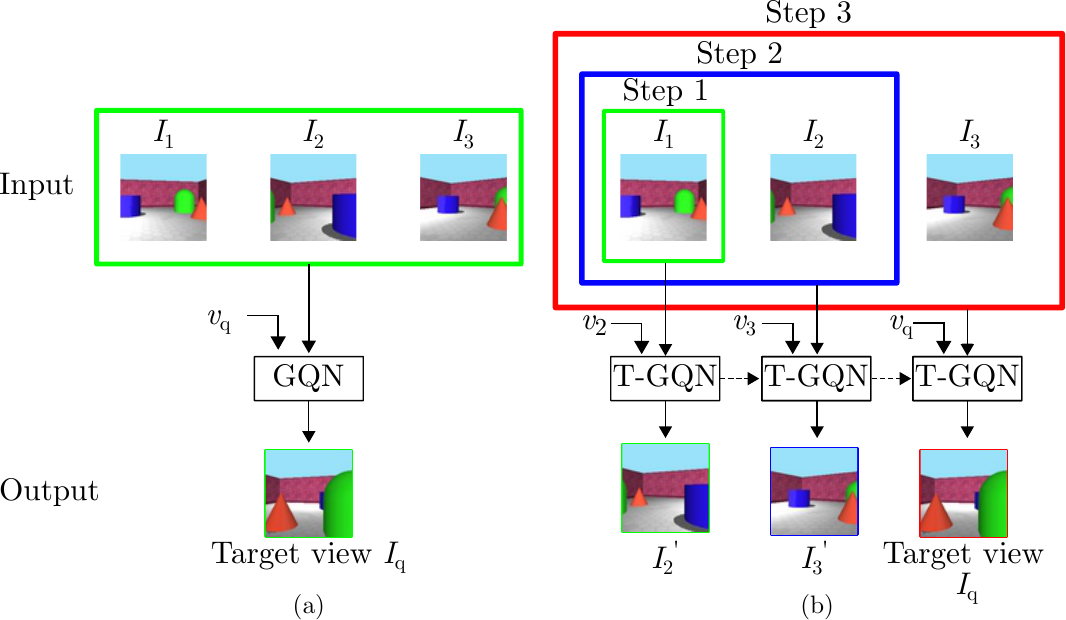}
    \caption{An illustration of the single view synthesis (a) compared to our proposed sequential view synthesis (b). Previously proposed GQN \cite{GQN} directly predicts the target view in a single rendering step, while our proposed T-GQN sequentially predicts novel views from an ordered set of $N$ observations. The output set includes $N-1$ nearest views of the target view and the target view itself. Colored boxes indicate which input observations are encoded at each rendering step.}
    \label{overall_seq}
\end{figure}

As can be seen from Fig.~\ref{overall_seq} (a), GQN \cite{GQN} predicts the target view $I_q$ in a single rendering step. If the query pose $v_q$ is distant from all input poses then the target view might look completely different than all input views. In this case, minimizing the above $\mathcal{L}_{GQN}$ loss does not guarantee to generate a plausible target view and it might take a long training time to reach the convergence.

We argue that if the model is able to predict an input view $I_n^\prime$ for $n>1$  based on previous input data $\{(I_1,v_1),\ldots, (I_{n-1},v_{n-1})\}$ then it also renders the target view $I_q$ at the query pose $v_q$ provided that the camera poses $\{v_1,\ldots, v_{N}, v_q\}$ have been organized as a sequence where the adjacent poses are the closest ones.
To achieve such ability, we train our proposed T-GQN model using multiple rendering steps. Each rendering step of our model is identical with GQN except that we use different sets of input observations and query poses. In Fig.~\ref{overall_seq}~(b) we illustrate these sets of input observations at each rendering step with boxes of different colors.

For example, in the first rendering step, we only allow the model to use the information inside the green box which includes the input view $I_1$ and its camera pose $v_1$. Then, our T-GQN model is trained to predict the next input view $I_2^\prime$ at the viewpoint $v_2$. Preventing the model to look at the input view $I_2$ would encourage the model to reason about the given contexts and produce a global implicit scene representation that extrapolates beyond the given information. In the final rendering step, both GQN and our proposed T-GQN model are allowed to see all $N$ input observations to render the target view but our model can leverage the past experiences from previous rendering steps to have a better approximation $I_q^\prime$ of the target image $I_q$. Furthermore, training our model in multiple rendering steps enforces our model to make consistent predictions of novel views at different viewpoints in a forward pass. This helps to stabilize the training process by making the network to produce deterministic results and also be able to reach the convergence faster than the former architecture. Note that our only use sequential view synthesis during training. In the testing time, we use randomly ordered sets of context views to have a fair comparison with other methods.

Therefore, the problem of single view synthesis can be redefined as the problem of sequential view synthesis by predicting a sequence of $N$ novel views $S_{out}=\{I_2^\prime,...,I_N^\prime,I_q^\prime\}$ from a sequence of observations $S_{in}=\{(I_1,v_1),...,(I_N,v_N)\}$. Since we are having $N$ different target views then it would be beneficial to have $N$ different scene representations. In order to have such multiple implicit scene representations, we use the Transformer Encoder \cite{Transformer_1} to learn the dependencies between input observations at each rendering step.

\subsection{Multi-view attention learning via Transformer Encoder}\label{multi-view_attention}
\begin{figure}[t]
    \centering
    \includegraphics[width=\textwidth]{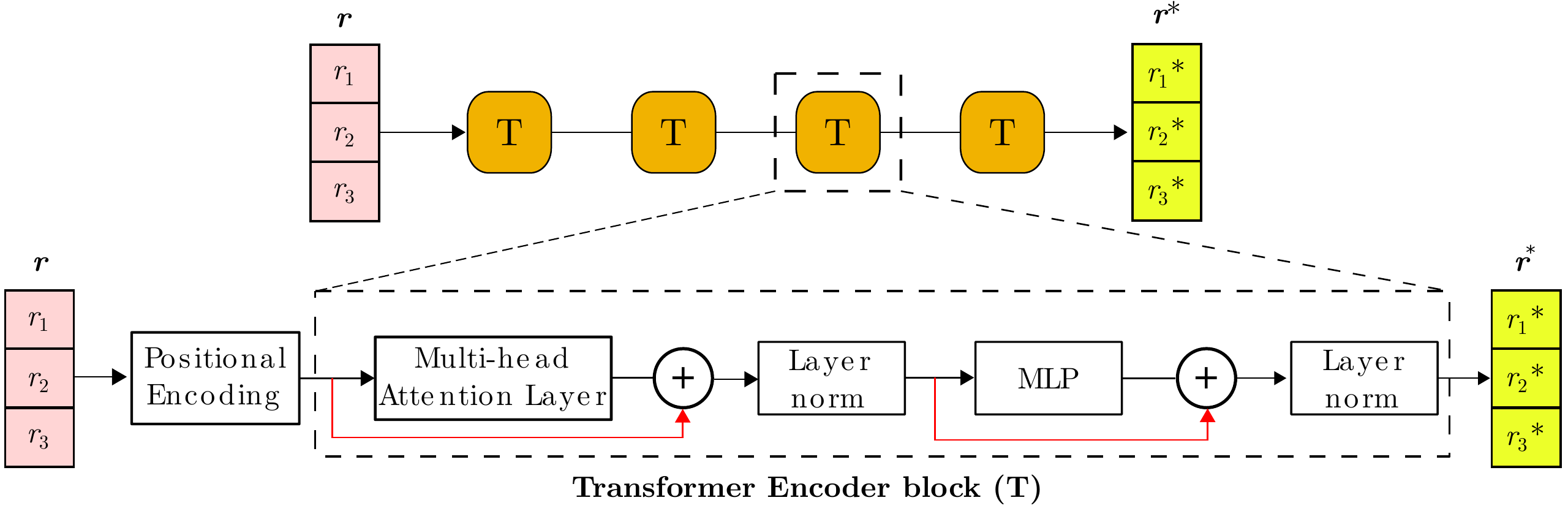}
    \caption{Illustration of multi-view attention learning using a stack of Transformer Encoder (T) blocks. For the visualization purposes, we show how a single Transformer Encoder block is able to produce multiple implicit scene representations $r^*$ using the input view representations $r$.}
    \label{fig_T_GQN}
\end{figure}

Recent works in the Language Modeling task \cite{Transformer_1,Transformer_2,Transformer_3,transformerXL} use the self attention-based Transformer Encoder to effectively learn dependencies between word embeddings in a sentence. Transformer Encoder takes a set of word embeddings as an input and produces another set of enhanced word embedding representations. Each of these representations reflects the long-range dependencies between input embeddings and they have been proven to be useful for training various language modeling tasks. Therefore, we can also take a set of scene representations $r$ as an input and use the Transformer Encoder to produce another set of enhanced scene representations $r^*$ that are trained to exploit the multi-view dependencies.

Fig.~\ref{fig_T_GQN} shows an example of how we use a stack of  Transformer Encoder blocks to perform the multi-view attention learning and implicitly represent a scene using a set of multiple representations. Within a Transformer Encoder block, the most important component is the Multi-Head Attention layer using the self-attention mechanism \cite{Transformer_1}. The self-attention function produces an $N \times N$ matrix $A$ of multi-view attention scores so that each row of the matrix $A$ reflects the learned multi-view dependencies at each rendering step. In addition, we apply an attention mask $m$ to the multi-view attention scores $A$. This attention mask allows us to control which part of the input representation sequence $r$ we would like the model to ignore when computing the attention scores. We leave the implementation details of Transformer Encoder to the supplementary material. In practice, we found that applying this attention mask leads to better performance on datasets that have high similarities between viewpoints. When the input viewpoints are not overlapping, we found that better results are achieved without applying the attention mask. 

\begin{figure}[!t]
    \centering
    \includegraphics[width=0.9\textwidth]{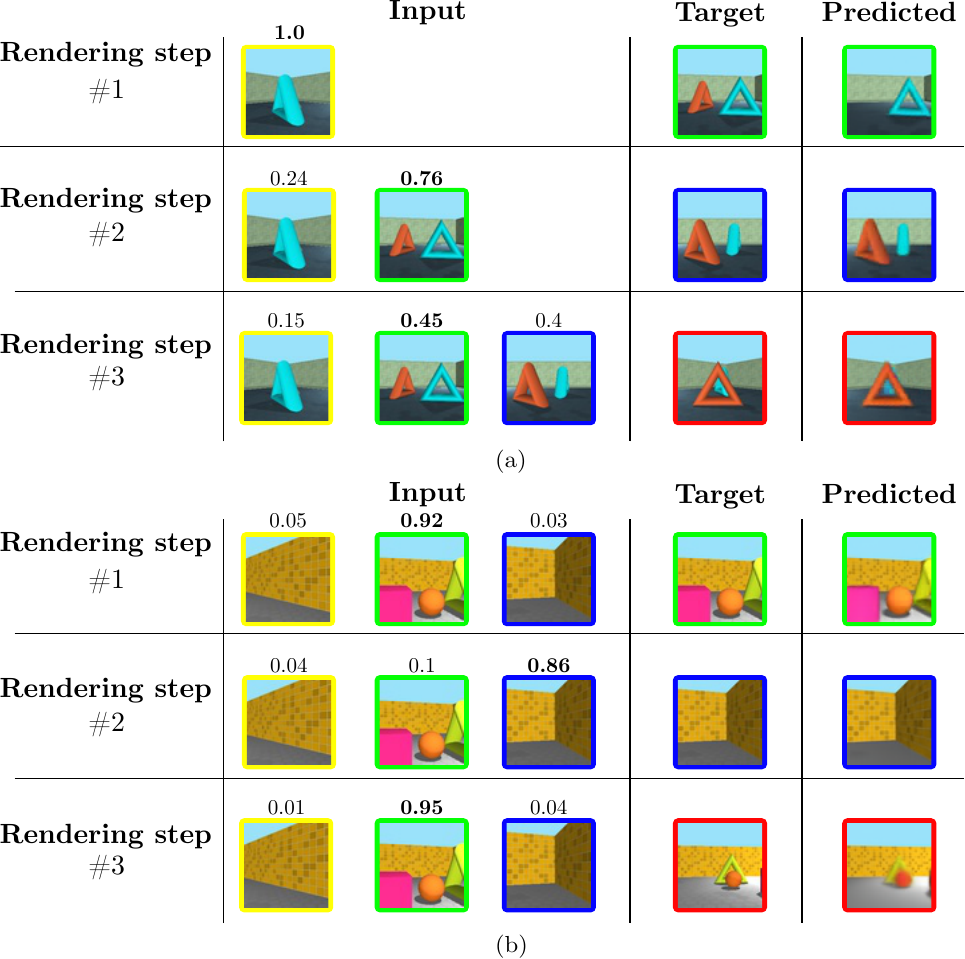} 
    \caption{Visualization of multi-view attention scores at each rendering step produced by our methods when (a) the camera movement is restricted and (b) the camera is free to move. Note that the order of the input sequence does not affect the learned multi-view attention scores.} 
    \label{fig_scores}
\end{figure}

If the movement of the camera is restricted then input viewpoints overlap with each other. Therefore, masking the attention scores would mean that the model is trained to predict the novel view using only a subset of the input sequence. Fig.~\ref{fig_scores} (a) shows the multi-view attention scores of a camera-restricted example. In the $1^{st}$ rendering step, we try to render the $2^{nd}$ input view (green box) as the novel view. Since our model is only allowed to use the $1^{st}$ view (yellow box) as the input, the predicted image only contains the cyan object. However, our model is able to render the missing orange object by putting higher scores on the $2^{nd}$ input view in the next rendering step. Since the $1^{st}$ input view does not have enough information to render the target view, our model learns to give it less attention and produce higher attention scores on another two input views. 

When the camera movement is not restricted, the input sequence might contain views which do not necessary include information to render the target view. In this case, we allow the model to use all input observations by not applying the attention mask. In Fig.~\ref{fig_scores}~(b), our model is able to give high attention scores to input observations which are the ground-truth views in the first two rendering steps. Therefore, the first two predicted novel views are identical to the ground-truth because the model has already seen these novel views as inputs. However, our goal is to be able to predict the target view (red box) which has a distant query pose. Among three input observations, the $2^{nd}$ input is the only view which contains information about objects on the scene. As can be seen in the last rendering steps, our model manages to pay most attention on the $2^{nd}$ input view (green box) and gives low scores to the other two views and successfully predicts the target view despite the distant query pose.

\subsection{Sequential rendering decoder}\label{sequentialDecoder}
\begin{figure}[!t]
    \centering
    \includegraphics[width=\textwidth]{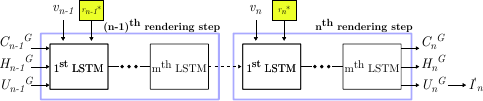}
    \caption{Illustration of our proposed sequential rendering decoder. At each rendering step, we use the computed states from the previous rendering step as the initialization for the decoder network. The canvas state $U_n^G$ is utilized to render the novel image $I_n^\prime.$ } 
    \label{fig_seq_decoder}
\end{figure}
In this paper, we improve the decoder network of GQN \cite{GQN} by applying a recurrent mechanism between rendering steps. In order to transfer the knowledge between rendering steps, computed LSTM states from previous steps are utilized as initialization for the current step as shown in the Fig.~\ref{fig_seq_decoder}. Here, we simplify the notation by defining that $C^G_n$ is the cell state of the $m^{th}$ Generation core at the $n^{th}$ rendering step. We apply this kind of notation to all cell states $C$, hidden states $H$ and canvas states $U$ of the decoder network. At the $n^{th}$ rendering step, the novel image $I_n^\prime$ is computed as follows:

\begin{equation}
    \label{equation_output}
    I_n^\prime = \text{decoder}(r_n^*,v_n,C_{n-1}^G,H_{n-1}^G,U_{n-1}^G,C_{n-1}^I,H_{n-1}^I)
\end{equation}

Since our model has multiple rendering steps, the prior and the posterior terms, $\pi^m_n$ and $q^m_n$, are obtained from the $m^{th}$ LSTM core at the $n^{th}$ rendering step. We train our model using the modified ELBO loss function as follows:

\begin{equation}
    \label{eqn_TGQN}
    \mathcal{L}_{TGQN} = \bigg[
    -\sum_{n=1}^N\ln\mathcal{N}(I_n|I_n^\prime)+\beta\sum_{n=1}^N\sum_{m=1}^{M} \text{KL}\Big[\mathcal{N}(q_n^m) || \mathcal{N}({\pi}_n^m)\Big]
\bigg]
\end{equation}
By adding the $\beta$ coefficient to the KL term, we emphasize discovering the disentangled latent factors \cite{VAE}. This technique has proven to be effective to maximize the probability of generating the desired output and keeping the distance between the prior and posterior distribution small \cite{betaVAE}. 

\section{Experiments}\label{Experiments}
\subsection{Experimental setup}
As illustrated in the Fig.~\ref{overall_seq}, 
to create a sequence of adjacent poses, we could, for example, use the overlap between the viewing frustums of the cameras to measure their adjacency. However, in our experiments we simply reordered all input observations based on the Euclidean distance between the translation vector of the query pose and the input poses that turned out to be sufficient to demonstrate the efficiency of our method. We choose to train our T-GQN model using three input observations to render a target view at an arbitrary query pose. Further details about our architecture are presented in the supplementary material.

\subsection{Comparing with state-of-the-art methods}

To evaluate our model, we compare against the GQN \cite{GQN} and an improved variant of E-GQN \cite{EGQN}. We use two datasets: Rooms-Ring-Camera (RRC) and Rooms-Free-Camera (RFC), from \cite{GQN} and one dataset: Rooms-Random-Objects (RRO) from \cite{EGQN}. The RRC dataset contains 2 millions rendered 3D square rooms that are composed of random objects of various shapes, colors and locations. Moreover, the scene textures, walls and lights are also randomly generated. In this dataset, the camera only moves on a fixed ring and always faces the center of the room. In case of the RFC dataset, the environment is the same with RRC except for the freely moving camera and objects rotated around their vertical axes. We also evaluate our method on the RRO dataset which includes complex 3D scenes with realistic 3D objects from the ShapeNet dataset \cite{ShapeNet}.

Recent works \cite{nerf,Plenoptic} on view synthesis often require retraining for any testing scene. Although, they produce visually impressive novel views but they are not able to generalize to unseen data. Therefore, we decide to compare our method against Scene Representation Network (SRN) \cite{srn} which is the current top-performing technique for view synthesis which generalizes reasonably well on testing data. Due to computational constraints, we can not compare this method with the full RRC or RFC dataset. We instead use a small subset of 1000 scenes from the Shapard-Metzler-7-Parts (SM7) and 3500 scenes from the RRC dataset to train both our proposed model and SRN model. In this experiment, the ratio between the training set and testing set is 9:1.

\begin{figure}[!t]
    \centering
    \includegraphics[width=\textwidth]{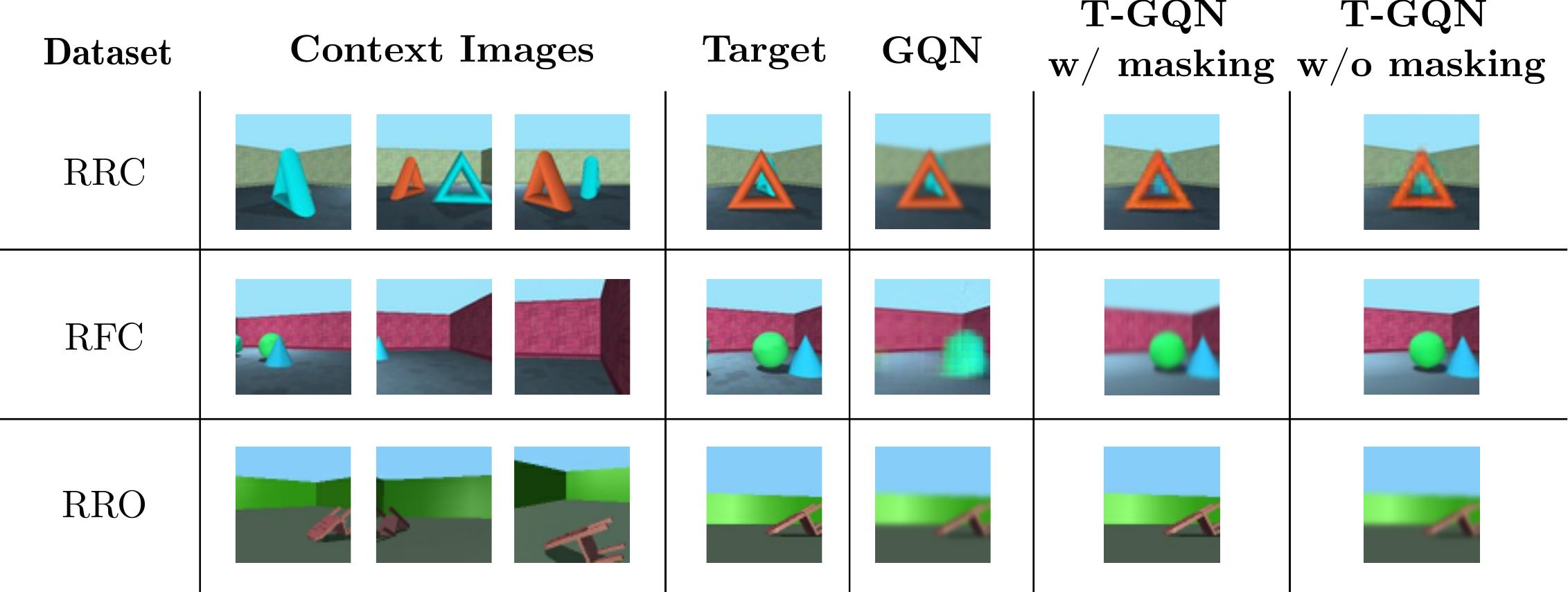}
    \caption{Example of rendered novel views using our proposed T-GQN (with and without masking) and other methods. Overall, our proposed T-GQN model is able to produce better target view than previous methods. More examples can be found in the Appendix section. We urge the reader to zoom in for better visualization.}
    \label{fig_overall_result}
\end{figure}

\begin{table}[!t]
\centering
\caption{Quantitative comparison of results between our T-GQN model, original GQN\cite{GQN} and E-GQN\cite{EGQN}.}
\label{Quant_table}
\resizebox{\textwidth}{!}{%
\begin{tabular}{@{}cccccccc@{}}
\toprule
\multirow{2}{*}{Model} &
  \multirow{2}{*}{\begin{tabular}[c]{@{}c@{}}\# parameters\\ (millions)\end{tabular}} &
  \multicolumn{3}{c}{L1 (pixels)} &
  \multicolumn{3}{c}{L2 (pixels)} \\ \cmidrule(l){3-8} 
                                                                &     & RRC         & RFC           & RRO          & RRC           & RFC           & RRO           \\ \midrule
\begin{tabular}[c]{@{}c@{}}GQN \\ (8 LSTM layers)\end{tabular}  & 381 & 8.50 $\pm$ 7.01  & 14.23 $\pm$ 13.56 & 14.12 $\pm$ 8.19 & 18.62 $\pm$ 12.86 & 30.28 $\pm$ 25.69 & 21.79 $\pm$ 10.23 \\
\begin{tabular}[c]{@{}c@{}}GQN \\ (12 LSTM layers)\end{tabular} & 428 & 7.40 $\pm$ 6.22 & 12.44 $\pm$ 12.89 & 10.12 $\pm$ 5.15 & 14.62 $\pm$ 12.77 & 26.80 $\pm$ 21.35 & 19.63 $\pm$ 9.14  \\
E-GQN                                                           & N/A & 3.59 $\pm$ 2.10 & 12.05 $\pm$ 11.79 & 6.59 $\pm$ 3.23  & 6.80 $\pm$ 5.23   & 27.65 $\pm$ 20.72 & 12.08 $\pm$ 6.52  \\
\begin{tabular}[c]{@{}c@{}}T-GQN\\ without mask\end{tabular} &
  382 &
  3.30 $\pm$ 2.05 &
  \textbf{9.25 $\pm$ 9.15} &
  6.31 $\pm$ 2.59 &
  6.65 $\pm$ 4.29 &
  \textbf{12.72 $\pm$ 10.56} &
  11.72 $\pm$ 5.23 \\
\begin{tabular}[c]{@{}c@{}}T-GQN\\ with mask\end{tabular} &
  382 &
  \textbf{2.31 $\pm$ 1.89} &
  11.65 $\pm$ 10.25 &
  \textbf{5.28 $\pm$ 2.31} &
  \textbf{5.92 $\pm$ 2.69} &
  15.44 $\pm$ 13.37 &
  \textbf{11.65 $\pm$ 5.13} \\ \bottomrule
\end{tabular}%
}
\end{table}

\subsection{Results}\label{results}

{\textbf{\textit{Comparing against GQN\cite{GQN} and E-GQN\cite{EGQN}:}}} In our experiment, we have tested our model to sample several novel views given the same input views from the same scene. In the Table \ref{Quant_table}, we have reported the L1 and L2 distances between the rendered novel views and the ground-truth images across testing scenes. Since our testing models are probabilistic, we run our experiment several times and report the average and standard deviation based on the test runs. In each testing scene, we sample 10 independent viewpoints and use 3 of them as the input to our model to generate the rest of viewpoints as novel images.\par

Qualitative results shown in Fig.~\ref{fig_overall_result} demonstrate that the T-GQN model with or without the attention mask is able to render all novel images much sharper than the former architectures. Meanwhile, predicted target views from GQN \cite{GQN} are often blurry and not able to get the correct object types, colors and positions. As explained in the Section \ref{seq_synthesis}, our model is able to predict nearby views of the novel view. Therefore, we include step-by-step renderings in the supplementary material. \par

As can be seen from Table \ref{Quant_table}, training our T-GQN model with the attention mask leads to a performance gain on the RRC and RRO datasets. In the case of the RFC dataset, our method without the attention mask performs significantly better than other methods including ours with the attention mask. In the free-camera case, the query poses are often far from the input poses so there may be not enough information from input views to generate the novel view. This explains the small gap of performance between our T-GQN model with masking on the RRC dataset and our T-GQN model without masking on the RFC dataset. In our paper, we apply a zero attention mask in testing the RFC dataset so that the model has access to all possible information. Our model then learns to attend to the relevant input views via our proposed multi-view attention learning described in Section \ref{multi-view_attention}. This highlights the strength of the Transformer architecture that learns to attend to the most relevant piece of information among input views.


\begin{figure}[!t]
    \centering
    \includegraphics[width=\textwidth]{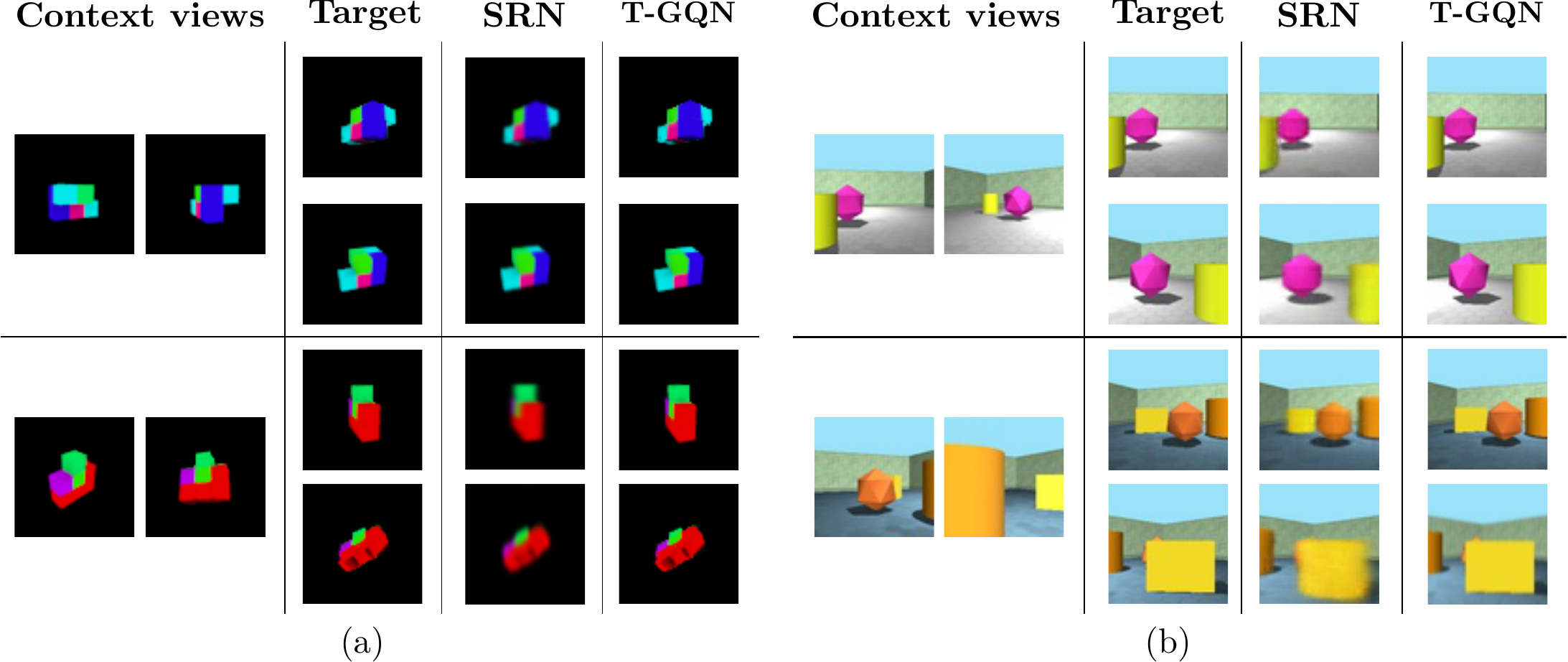}
    \caption{Qualitative results between our proposed T-GQN and SRN on (a) SM7 dataset and (b) RRC dataset.}
    \label{qual_srn} 
\end{figure}

\begin{table}[!t]
\centering
\caption{Quantitative comparison of results between SRN, our proposed T-GQN model tested on the small subset of SM7 and RRC dataset.}
\label{quant_srn}
\begin{tabular}{@{}lcccc@{}}
\toprule
\multirow{2}{*}{Method} & \multicolumn{2}{c}{L1 (pixels)}                          & \multicolumn{2}{c}{L2 (pixels)}            \\ \cmidrule(l){2-5} 
                        & SM7                           & RRC             & SM7             & RRC             \\ \midrule
SRN                     & 2.15                          & 3.56            & 5.65            & 6.21            \\
T-GQN                   & 2.14 $\pm$ 0.21 & 1.92 $\pm$ 0.15 & 5.25 $\pm$ 0.34 & 2.96 $\pm$ 0.93 \\ \bottomrule
\end{tabular}
\end{table}

\noindent\textbf{\textit{Comparing against SRN\cite{srn}:}}
We also benchmark novel view synthesis accuracy on few-shot reconstruction to compare our probabilistic T-GQN model with a deterministic view synthesis SRN \cite{srn}. As can be seen in Table.~\ref{quant_srn} and Fig.~\ref{qual_srn}, our proposed method manages to achieve similar results with SRN model on the SM7 dataset. In case of RRC dataset, our model performs significantly better than SRN. We argue that SM7 is an easy dataset which contains only a single rotating colored object. Therefore, learning to generalize between training and testing data of SM7 dataset is easier than with the RRC dataset. 

As the number of training scenes increase, training a deterministic approach such as SRN to have good generalization across a large number of testing scenes is difficult and still an open challenge. Qualitative results in Fig.~\ref{qual_srn} (b) shows that our T-GQN model is not only able to render correct room layouts and object properties but also produce significantly sharper results than the previously proposed SRN method with the demanding RRC dataset. We discuss more about the capabilities of rendering novel views between our method and SRN in the supplementary material.

\section{Ablation Study}

\begin{figure}[!t]
    \centering
    \includegraphics[width=0.9\textwidth]{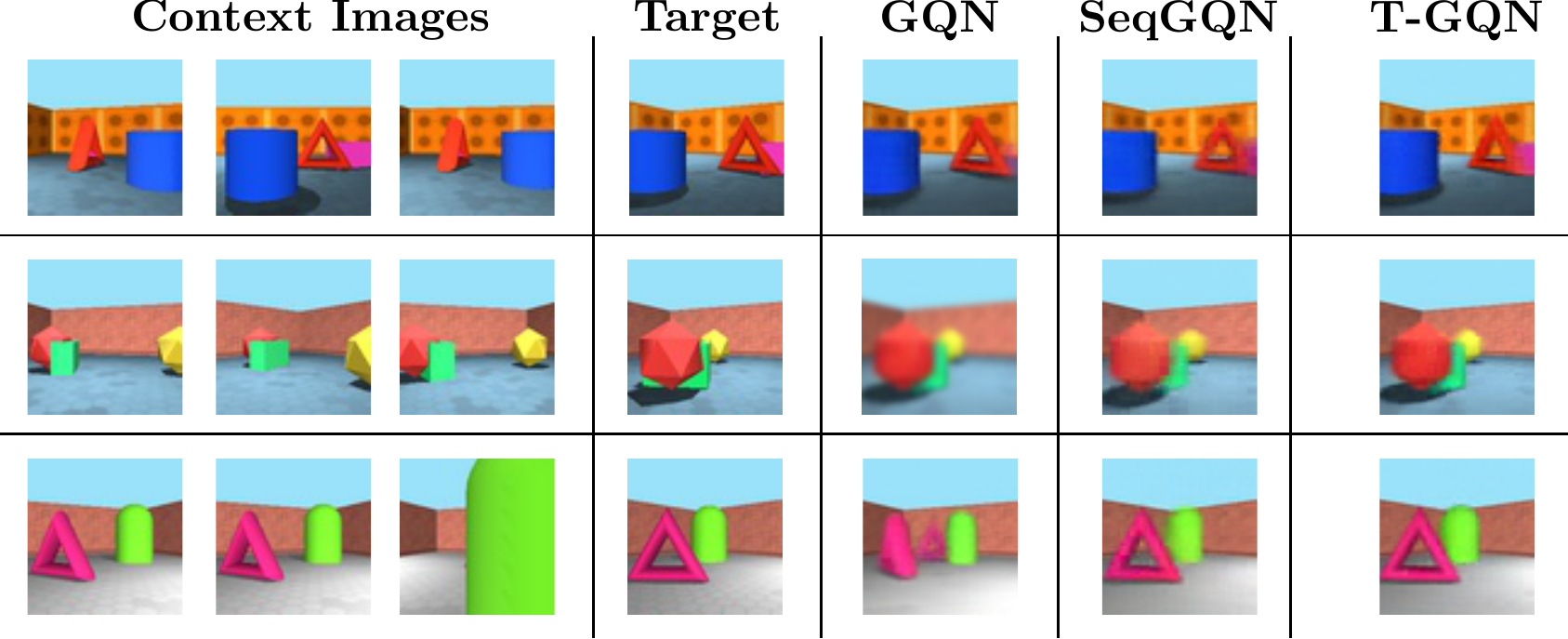}
    \caption{Example of generated novel views compared between our proposed T-GQN and variants on the RRC dataset.}
    \label{fig_SegvST}
\end{figure}

\begin{table}[!t]
\centering
\caption{Quantitative results of T-GQN and its variants on the Rooms-Ring-Camera dataset.}
\label{table_seq_vs_T}
\begin{tabular}{@{}ccccc@{}}
\toprule
Metric & GQN & SeqGQN & \begin{tabular}[c]{@{}c@{}}T-GQN\\ ($\beta = 1$)\end{tabular} & \begin{tabular}[c]{@{}c@{}}T-GQN\\ ($\beta = 250$)\end{tabular} \\ \midrule
L1 (pixels) & 7.40 $\pm$ 6.22  & 4.51 $\pm$ 2.35  & 3.7 $\pm$ 1.92   & \textbf{2.31 $\pm$ 1.89} \\
L2 (pixels) & 14.62 $\pm$ 12.77 & 8.2 $\pm$ 4.27  & 7.1 $\pm$ 2.47  & \textbf{5.92 $\pm$ 2.69} \\
SSIM        & 0.85  & 0.882 & 0.905 & \textbf{0.92} \\ \bottomrule

\end{tabular}
\end{table}

To investigate the effectiveness of the proposed modules, we use our sequential rendering decoder to generate novel views using the aggregated scene representation $R$ from \cite{GQN} and denote this model as SeqGQN. Instead of using different multi-view attention representations at each rendering step, we use a single representation $R$ as an input to all ConvLSTM layers in our proposed sequential rendering decoder network. The qualitative and quantitative results including the rendered novel views using SeqGQN are shown in Fig.~\ref{fig_SegvST} and Table \ref{table_seq_vs_T}. We observe that our full model significantly outperforms both SeqGQN and GQN. Although both GQN and SeqGQN are using the same aggregated scene representation $R$, SeqGQN is able to produce better and more accurate target views than the baseline. This result demonstrates that approaching the neural rendering in the sequential manner leads to more accurate view synthesis. We also find that each learned representation is fully capable of rending the novel view. Therefore, we can use a single scene representation $r^*_n$ to render the novel view in the testing time. Further explanations are included in the supplementary material.

Since we are adding a large $\beta$ coefficient into the KL loss term (Eq.~\ref{eqn_TGQN}), we are encouraging the model to approximate the posterior distribution $q^m_n$ close to the prior distribution $\pi^m_n$ as much as possible.  If the query pose is far away from given input observations then it is hard to get a good estimate for $q^m_n$. In \cite{GQN}, the model is prone to mistakes because the target view is rendered in a single step. Using the sequential rendering decoder, we solve the problem by generating the nearby views of the target view before rendering the target view. As can be seen from Table \ref{table_seq_vs_T}, training our T-GQN model with $\beta = 1$ leads to a drop of the overall performance. However, this model is still able to outperform both GQN and SeqGQN. 

\begin{figure}[t!]
    \centering
    \includegraphics[width=0.9\textwidth]{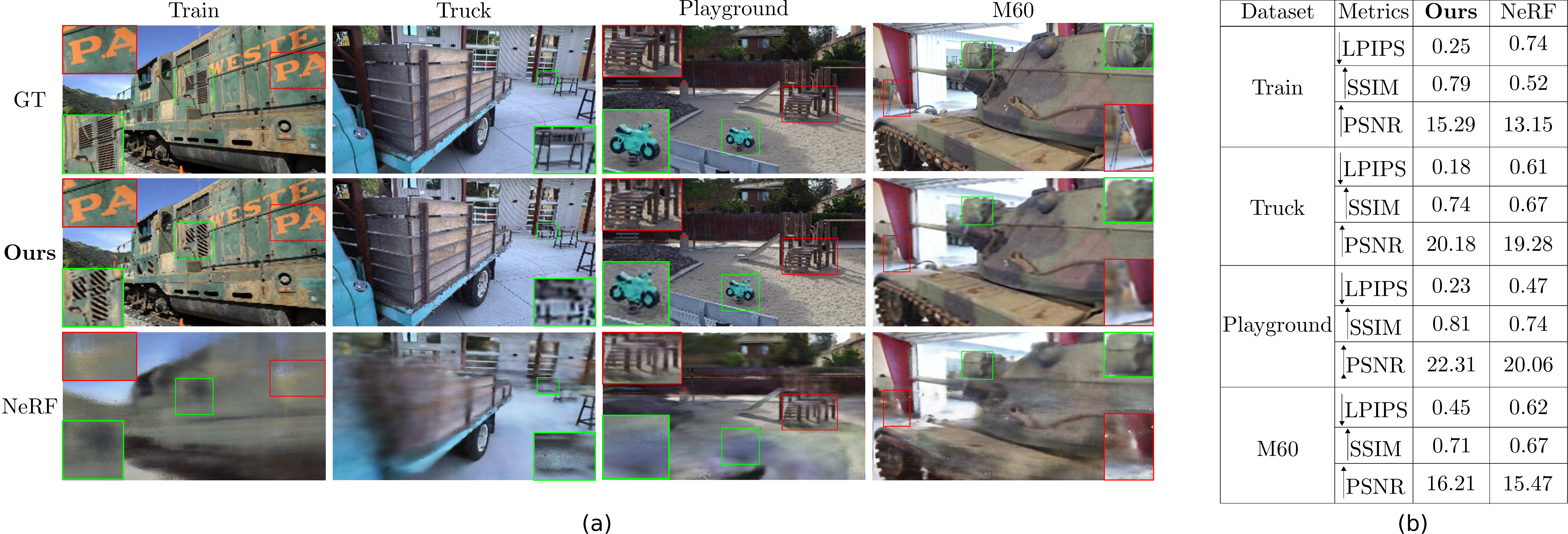}
    \caption{Qualitative(a) and quantitative(b) results on 4 testing scenes of Tank and Temples dataset.}
    \label{NeRF}
\end{figure}

To evaluate our method on the real data, we compare our method with the recently proposed NeRF \cite{nerf} on the challenging Tanks and Temples dataset. We train our proposed method on 10 randomly selected training scenes and evaluate the performance on 4 unseen testing scenes. Note that our method haven't seen any testing images during training or require any retraining like NeRF. However, as can be seen in Fig.~\ref{NeRF}, our method outperforms NeRF both qualitatively and quantitatively. This may be explained by the fact that NeRF makes an assumption of inward-facing scenes while our method does not have such limitations. On the flip side, our proposed method is not bounded to this condition and can generate plausible novel views in a wide variety of 3D scenes. This highlights the effectiveness of our method that generalize reasonably well on unseen data.

\section{Conclusions}

In this paper, we presented a method to synthesize novel views in a sequential manner. Instead of directly rendering the target view, we train our model to predict a sequence of novel views in multiple rendering steps. Using the Transformer Encoder, our proposed multi-view attention learning is able to learn different implicit scene representations for each rendering step. The experimental results demonstrate that our model is able to render more accurate novel views using less training time.

In the future works, we will explore different implicit neural scene representations to further improve the quality of synthesized images. Another interesting approach would be learning to synthesize novel views that are useful for further tasks such as 3D scene reconstruction. 

\bibliographystyle{splncs}
\bibliography{egbib}

\end{document}